\documentclass[runningheads,a4paper]{llncs}

\usepackage{amssymb}
\usepackage{amsmath}
\usepackage{graphicx}
\usepackage{multirow}
\usepackage[normalem]{ulem}
\useunder{\uline}{\ul}{}
\usepackage{bbding}
\usepackage{adjustbox}

\usepackage{pbox}
\usepackage{makecell}
\usepackage{ifxetex}
\usepackage{ifluatex}

\ifxetex
  \usepackage[ios,font=seguiemj.ttf]{emoji}
  \usepackage{fontspec}
  
\else
  \ifluatex
    \usepackage[ios,font=Symbola_hint.ttf]{emoji}
    \usepackage{fontspec}
    
  \else
    \usepackage[T1]{fontenc}
    \usepackage[utf8]{inputenc}
    \usepackage[ios]{emoji}
  \fi
\fi

\usepackage{url}
\urldef{\mailsa}\path|{sanjaya, lakshika, amit, derek}@knoesis.org|
\newcommand{\keywords}[1]{\par\addvspace\baselineskip
\noindent\keywordname\enspace\ignorespaces#1}

\begin{document}

\mainmatter  

\pdfinfo{
   /Author (Sanjaya Wijeratne, Lakshika Balasuriya, Amit Sheth, Derek Doran)
   /Title (A Semantics-Based Measure of Emoji Similarity)
   /Keywords (Emoji Similarity; Emoji Analysis and Search; Semantic Similarity)
   /Subject (A Semantics-Based Measure of Emoji Similarity - Web Intelligence 2017)
}

\title{A Semantics-Based Measure of Emoji Similarity}

\titlerunning{A Semantics-Based Measure of Emoji Similarity}

\author{Sanjaya Wijeratne \and Lakshika Balasuriya \and Amit Sheth \and Derek Doran}
\authorrunning{Wijeratne et al.}

\institute{Kno.e.sis Center, Wright State University\\
Dayton, Ohio, USA\\
\mailsa\\
\url{http://www.knoesis.org}}

\maketitle

\begin{abstract}

Emoji have grown to become one of the most important forms of communication on the web. With its widespread use, measuring the similarity of emoji has become an important problem for contemporary text processing since it lies at the heart of sentiment analysis, search, and interface design tasks. This paper presents a comprehensive analysis of the \emph{semantic} similarity of emoji through embedding models that are learned over machine-readable emoji meanings in the EmojiNet knowledge base. Using emoji descriptions, emoji sense labels and emoji sense definitions, and with different training corpora obtained from Twitter and Google News, we develop and test multiple embedding models to measure emoji similarity. To evaluate our work, we create a new dataset called EmoSim508, which assigns human-annotated semantic similarity scores to a set of 508 carefully selected emoji pairs. After validation with EmoSim508, we present a real-world use-case of our emoji embedding models using a sentiment analysis task and show that our models outperform the previous best-performing emoji embedding model on this task. The EmoSim508 dataset and our emoji embedding models are publicly released with this paper and can be downloaded from \url{http://emojinet.knoesis.org/}.


\end{abstract}

\section{Introduction}

With the rise of social media, pictographs, commonly referred to as `emoji' have become one of the world's fastest-growing forms of communication\footnote{\url{https://goo.gl/jbeRYW}}. This rapid growth of emoji began in 2011 when the Apple iPhone added an emoji keyboard to iOS, and again in 2013 when the Android mobile platform started to support emoji on their mobile devices~\cite{instagramstudy}. Emoji permeate modern online and web-based communication and are now regarded as a natural and common form of expression. In fact, the Oxford Dictionary named `face with tears of joy' \emoji{1F602} as the word of the year in 2015\footnote{\url{https://goo.gl/6oRkVg}}. Not only individuals but also business organizations have adopted emoji with a 777\% year-over-year increase and 20\% month-over-month increase in emoji usage for marketing campaigns in 2016\footnote{\url{https://goo.gl/ttxyP1}}. Major search engines, including Bing\footnote{\url{https://goo.gl/5iy8Dx}} and Google\footnote{\url{https://goo.gl/oDfZTQ}}, now support web searches involving emoji as search terms.

As analysis and modeling of written text by Natural Language Processing (NLP) techniques have enabled important advances such as machine translation~\cite{weaver1955translation}, word sense disambiguation~\cite{navigli2009word}, and search~\cite{guha2003semantic}, the transfer of such methods (or development of new methods) over emoji is only beginning to be explored~\cite{emojineticwsm}. The ability to automatically process, derive meaning, and interpret text fused with emoji will be essential as society embraces emoji as a standard form of online communication. Foundational to many emoji analysis tasks will be a way to measure  {\em similarity}, including:~(i) corpus searching, where documents (or a query) contains emoji symbols~\cite{cappallo2015query};~(ii) sentiment analysis~\cite{barbieri2016does,eisner2016emoji2vec}, where emoji sentiment lexicons~\cite{novak2015sentiment} are known to improve the performance; and~(iii) interface design, mainly in optimizing mobile phone keyboards~\cite{pohl2017beyond}. In fact, as of 2017, the poor design of emoji keyboards for mobile devices may be relatable to the reader: there are 2,389 emoji supported by the Unicode Consortium, yet listing and searching through all of them on a mobile keyboard is a time consuming task. Grouping similar emoji together could lead to optimized emoji keyboard designs for mobile devices~\cite{pohl2017beyond}.

The notion of the similarity of two emoji is very broad. One can imagine a similarity measure based on the pixel similarity of emoji pictographs, yet this may not be useful since the pictorial representation of an emoji varies by mobile and computer platform~\cite{miller2016blissfully,tigwell2016oh,cramer2016sender}. Two similar looking pictographs may also correspond to emoji with radically different senses (e.g., twelve thirty \emoji{1F567} and six o'clock \emoji{1F555}, raised hand \emoji{0270B} and raised back of hand \emoji{1F91A}, octopus \emoji{1F419}, and squid \emoji{1F991}, etc.)~\cite{emojinet,emojineticwsm}. Instead, we are interested in measuring the {\em {semantic}} similarity of emoji such that the measure reflects the {\em{likeness of their meaning, interpretation or intended use}}. Understanding the semantics of emoji requires access to a repository of emoji meanings and interpretations. The release of a new resource called {\em EmojiNet}~\cite{emojineticwsm} offers free and open access to an aggregation of such meanings and interpretations (called senses) collected from major emoji databases on the Internet (e.g., The Unicode Consortium, The Emoji Dictionary, and Emojipedia).

A collection of emoji sense definitions can enable a semantics-based measure of similarity through vector word embeddings. Word embeddings are a powerful and proven way~\cite{DBLP:journals/corr/abs-1301-3781} to measure word similarity based on their meaning. They have been widely used in semantic similarity tasks~\cite{hill2016simlex,huang2012improving,camacho2015nasari} and empirically shown to improve the performance of word similarity tasks when used with proper parameter settings~\cite{levy2015improving}. 
Word vectors also provide a convenient way of comparing them across each other. Thus, representing the emoji meanings using word embedding models can be used to generate word vectors that encode emoji meanings, which we call \textbf{emoji embedding models}.

In this paper, we present a comprehensive study on measuring the semantic similarity of emoji using emoji embedding models. We extract machine-readable emoji meanings from EmojiNet to model the meaning of an emoji. Using pre-trained word embedding models learned over a Twitter dataset of 110 million tweets and a Google News text corpus of 100 billion words, we encode the extracted emoji meanings to obtain emoji embedding models. To create a gold standard dataset for evaluating how well the emoji embeddings measure similarity, we ask ten human annotators who are knowledgeable about emoji to manually rate the similarity of 508 pairs of emoji. This dataset of human annotations, which we call `EmoSim508', is made available with this paper for use by other researchers. We evaluate the emoji embeddings by first establishing that the similarity measured by our embedding models align with the ratings of the human annotators using statistical measures. Then, we apply our emoji embedding models to a sentiment analysis task to demonstrate the utility of them in a real-world NLP application. Our models were able to correctly predict the sentiment class of tweets laden with emoji from a benchmark dataset~\cite{novak2015sentiment} with an accuracy of 63.6 (7.73\% improvement), outperforming the previous best results on the same dataset~\cite{barbieri2016does,eisner2016emoji2vec}.

This paper is organized as follows. Section~\ref{sec:rr} discusses the related literature and frames how this work differs from and furthers existing research. Section~\ref{sec:learnemojiembed} discusses how the emoji meanings are represented using the different emoji definitions extracted from EmojiNet and how the emoji embeddings are learned. Section~\ref{sec:dc} explains the creation of the EmoSim508 dataset. Section~\ref{sec:eval} reports how well the emoji embedding models perform on an emoji similarity analysis task and Section~\ref{sec:usecase} reports the performance of our emoji embedding models in a downstream sentiment analysis task. Section \ref{sec:con} offers concluding remarks and plans for future work.

\section{Related Work} \label{sec:rr}

While emoji were introduced in the late 1990s, their use and popularity was limited until the Unicode Consortium started to standardize emoji symbols in 2009~\cite{emojinet}. Major mobile phone manufactures such as Apple, Google, Microsoft, and Samsung then began supporting emoji in their device operating systems between 2011 and 2013, which boosted emoji adoption around the world~\cite{instagramstudy}. Early research on emoji was focused on understanding the role of emoji in computer-mediated communication. Kelly {\em{et al.}} studied how people in close relationships use emoji in their communications and reported that they use emoji as a way of making their conversations playful~\cite{kelly2015characterising}. Pavalanathan {\em{et al.}} studied how Twitter users adopt emoji and reported that Twitter users prefer emoji over emoticons~\cite{pavalanathan2016more}. Researchers have also studied how emoji usage and interpretation differ across mobile and computer platforms~\cite{miller2016blissfully,tigwell2016oh,cramer2016sender}, geographies~\cite{ljubevsic2016global}, and across languages~\cite{barbieri2016cosmopolitan} where many others have used emoji as features in their learning algorithms for problems such as emoji-based search~\cite{cappallo2015query}, sentiment analysis~\cite{novak2015sentiment}, emotion analysis~\cite{wang2012harnessing}, and Twitter profile classification~\cite{lakshikagang,gangwordembeddings}.

Emoji similarity has received little attention apart from three attempts by Barbieri {\em{et al.}}~\cite{barbieri2016does}, Eisner {\em{et al.}}~\cite{eisner2016emoji2vec} and Pohl {\em{et al.}}~\cite{pohl2017beyond}. Barbieri {\em{et al.}}~\cite{barbieri2016does} collected a sample of 10 million tweets originated from the USA and trained an emoji embedding model using tweets as the input. Then, using 50 manually-generated emoji pairs annotated by humans for emoji similarity and relatedness, they evaluated how well the learned emoji embeddings align with the human annotations. They reported that the learned emoji embeddings align more closely with the relatedness judgment scores of human annotators than the similarity judgement scores. Eisner {\em{et al.}}~\cite{eisner2016emoji2vec} used a word embedding model learned over the Google News corpus\footnote{\url{https://goo.gl/QaxjVC}}, applied it to emoji names and keywords extracted from the Unicode Consortium website, and learned an emoji embedding model which they called \texttt{emoji2vec}. Using t-SNE for data visualization~\cite{maaten2008visualizing}, Eisner {\em{et al.}} showed that the high dimensional emoji embeddings learned by \texttt{emoji2vec} could group emoji into clusters based on their similarity. They also showed that their emoji embedding model could outperform Barbieri {\em{et al.}}'s model in a sentiment analysis task. Pohl {\em{et al.}}~\cite{pohl2017beyond} studied the emoji similarity problem using two methods; one based on the emoji keywords extracted from the Unicode Consortium website and the other based on emoji embeddings learned from a Twitter message corpus. They used the Jaccard Coefficient\footnote{\url{https://goo.gl/RKkRzF}} on the emoji keywords extracted from the Unicode Consortium to find the similarity of two emoji. They evaluated their approach using 90 manually-generated emoji pairs and argued for how emoji similarity can be used to optimize the design of emoji keyboards.

Our work differs from the related works discussed above in many ways. Barbieri {\em{et al.}}~\cite{barbieri2016does} use the distributional semantics~\cite{harris1954distributional} of words learned over a Twitter corpus where they seek an understanding of emoji meanings from how emoji are used in a large collection of tweets. In contrast, this paper learns emoji embeddings based on emoji meanings extracted from EmojiNet. We learn the distributional semantics of the words in emoji definitions using word embeddings learned over two large text corpora and use the learned word embeddings to model the emoji meanings extracted from EmojiNet. Hence, we combine emoji meanings extracted from knowledge bases (i.e., EmojiNet) with distributional semantics of those words in emoji definitions. Pohl {\em{et al.}}~\cite{pohl2017beyond} learn emoji embedding models in the same way as Barbieri {\em{et al.}} and use the Jaccard Coefficient on emoji keywords extracted from the Unicode Consortium to measure similarity. This is similar to our earlier work on emoji similarity~\cite{emojineticwsm}, which we build upon in this paper. Eisner {\em{et al.}}'s~\cite{eisner2016emoji2vec} presented an embedding model built on short emoji names and keywords listed on the Unicode Consortium website, which is approximately 4 to 5 words long on average as reported by Pohl {\em{et al.}} in~\cite{pohl2017beyond}. Since prior research suggests that the emoji embedding models can be improved by incorporating more words by using longer emoji definitions~\cite{eisner2016emoji2vec,pohl2017beyond}, we introduce embeddings based on three different types of long-form definitions of an emoji.

\begin{table}[]
\centering
\caption{Nonuple Representation of an Emoji}
\label{emoji_representation}
\begin{tabular}{|l|l|}
\hline
\multicolumn{1}{|c|}{\textbf{Nonuple Element}} & \multicolumn{1}{c|}{\textbf{Description}}                                \\ \hline
Unicode $u_i$                            & U+1F64C                                                                  \\ \hline
Emoji Name $n_i$                              & Raising Hands                                                            \\ \hline
Short Code $c_i$                        & :raised\_hands:                                                          \\ \hline
Definition $d_i$                        & \pbox{5cm}{Two hands raised in the air, \\celebrating success or an event}.            \\ \hline
Keywords $K_i$                       & celebration, hand, hooray, raised                               \\ \hline
Images $I_i$                            & \emoji{1F64A} \emoji{1F64X} \emoji{1F64T} \emoji{1F64M} \emoji{1F64B} \emoji{1F64G}                                                                         \\ \hline
Related Emoji $R_i$                     & Confetti Ball, Clapping Hands Sign                             \\ \hline
Emoji Category $H_i$                    & Gesture symbols                                                          \\ \hline
Senses $S_i$           & \pbox{5cm}{Sense Label: celebration(Noun) \\ Def: A joyful occasion for special \\festivities to mark a happy event.} \\ \hline
\end{tabular}
\end{table}

\section{Learning Emoji Embedding Models} \label{sec:learnemojiembed}
In this section, we briefly present the EmojiNet resource and the different types 
of emoji sense definitions it contains. We subsequently discuss the training of emoji embedding models, constructed from the sense definitions extracted from EmojiNet. 

\subsection{EmojiNet}

EmojiNet is a comprehensive machine-readable emoji sense inventory~\cite{emojineticwsm}. EmojiNet maps emoji to their set of possible meanings or {\em senses}. It consists of 12,904 sense labels over 2,389 emoji, which were extracted from the web and linked to machine-readable sense definitions seen in BabelNet~\cite{navigli2010babelnet}. Each emoji in EmojiNet is represented as a nonuple representing its sense and other metadata. For each emoji $e_i$, the nonuple is given as $e_i = (u_i, n_i, c_i, d_i, K_i, I_i, R_i, H_i, S_i)$, where $u_i$ is the Unicode representation of $e_i$, $n_i$ is the name of $e_i$, $c_i$ is the short code of $e_i$, $d_i$ is a description of $e_i$, $K_i$ is the set of keywords that describe intended meanings attached to $e_i$, $I_i$ is the set of images that are used in different rendering platforms, $R_i$ is the set of related emoji extracted for $e_i$, $H_i$ is the set of categories that $e_i$ belongs to, and $S_i$ is the set of different senses in which $e_i$ can be used within a sentence. Apart from this, each sense $s_i \in S_i$ is defined as a combination of a word (e.g., \texttt{laugh}), its part-of-speech (POS) tag (e.g., \texttt{noun}), and its definition in a message context or gloss (e.g., \texttt{Produce laughter}). An example of the nonuple notation is shown in Table~\ref{emoji_representation}. EmojiNet is hosted as an open service with a REST API at \url{http://emojinet.knoesis.org/}.

\subsection{Representation of Emoji Meaning} \label{sec:emojirep}
We consider three different ways to represent the meaning of an emoji using the information in EmojiNet. Specifically, we extract emoji descriptions, emoji sense labels, and the emoji sense definitions of each emoji sense from EmojiNet to model the meaning of an emoji. We discuss each briefly below: 

\noindent{\textbf{Emoji Description {\em{(Sense\_Desc.)}}}:} Emoji descriptions give an over-view of what is depicted in an emoji and its intended uses. For example, for the pistol emoji \emoji{1F52B}, EmojiNet lists ``{\em{A gun emoji, more precisely a pistol. A weapon that has potential to cause great harm. Displayed facing right-to-left on all platforms}}'' as its description. One could use this information to get an understanding of how the pistol emoji should be used in a message. 

\noindent{\textbf{Emoji Sense Labels {\em{(Sense\_Label)}}}:} Emoji sense labels are word-POS tag pairs (such as \texttt{laugh(noun)}) that describe the senses and their part-of-speech under which an emoji can be used in a sentence. Emoji sense labels can act as words that convey the meaning of an emoji and thus, can play an important role in understanding the meaning of an emoji. For example, for pistol emoji \emoji{1F52B}, EmojiNet lists 12 emoji sense labels consisting of 6 nouns \texttt{(gun, weapon, pistol, violence, revolver, handgun)}, 3 verbs \texttt{(shoot, gun, pistol)} and 3 adjectives \texttt{(deadly, violent, deathly)}.

\noindent{\textbf{Emoji Sense Definitions {\em{(Sense\_Def.)}}}:} Emoji sense definitions are the textual descriptions that explain each sense label and how those sense labels should be used in a sentence. For example, for the \texttt{gun(Noun)} sense label of the pistol emoji \emoji{1F52B}, EmojiNet lists 5 sense definitions that complement each other\footnote{\url{https://goo.gl/gm7TQ2}}. These emoji sense definitions can be valuable in understanding the meaning of an emoji; thus, we use them to represent the meaning of an emoji.

\subsection{Learning the Emoji Embedding Models} \label{emojiembeddinglearning}

Once the machine-readable emoji descriptions are extracted from EmojiNet, we use word embedding models~\cite{DBLP:journals/corr/abs-1301-3781} to convert them into a vectorial representation. A word embedding model is a neural network that learns rich representations of words in a text corpus. It takes data from a large, $n$-dimensional `word space' (where $n$ is the number of unique words in a corpus) and learns a transformation of the data into a lower $k$-dimensional space of real numbers. This transformation is developed in a way that similarities between the $k$-dimensional vector representation of two words reflects semantic relationships among the words themselves. Word embedding models are inspired by the distributional hypothesis (i.e., words that are co-occurring in the same contexts tend to carry similar meanings), hence the semantic relationships among word vectors are learned based on the word co-occurrence in contexts (e.g., sentences) extracted from large text corpora. Mikolov {\em{et al.}} have shown that these word embeddings can learn different types of semantic relationships, including \texttt{gender relationships (e.g., King-Queen)} and \texttt{class inclusion (e.g., Clothing-Shirt)} among many others~\cite{mikolov2013linguistic}. Similar to word embedding models, an emoji embedding model is defined as an emoji symbol and its learned $k$-dimensional vector representation.

\begin{figure*}
\centering
\includegraphics[width=1.0\linewidth]{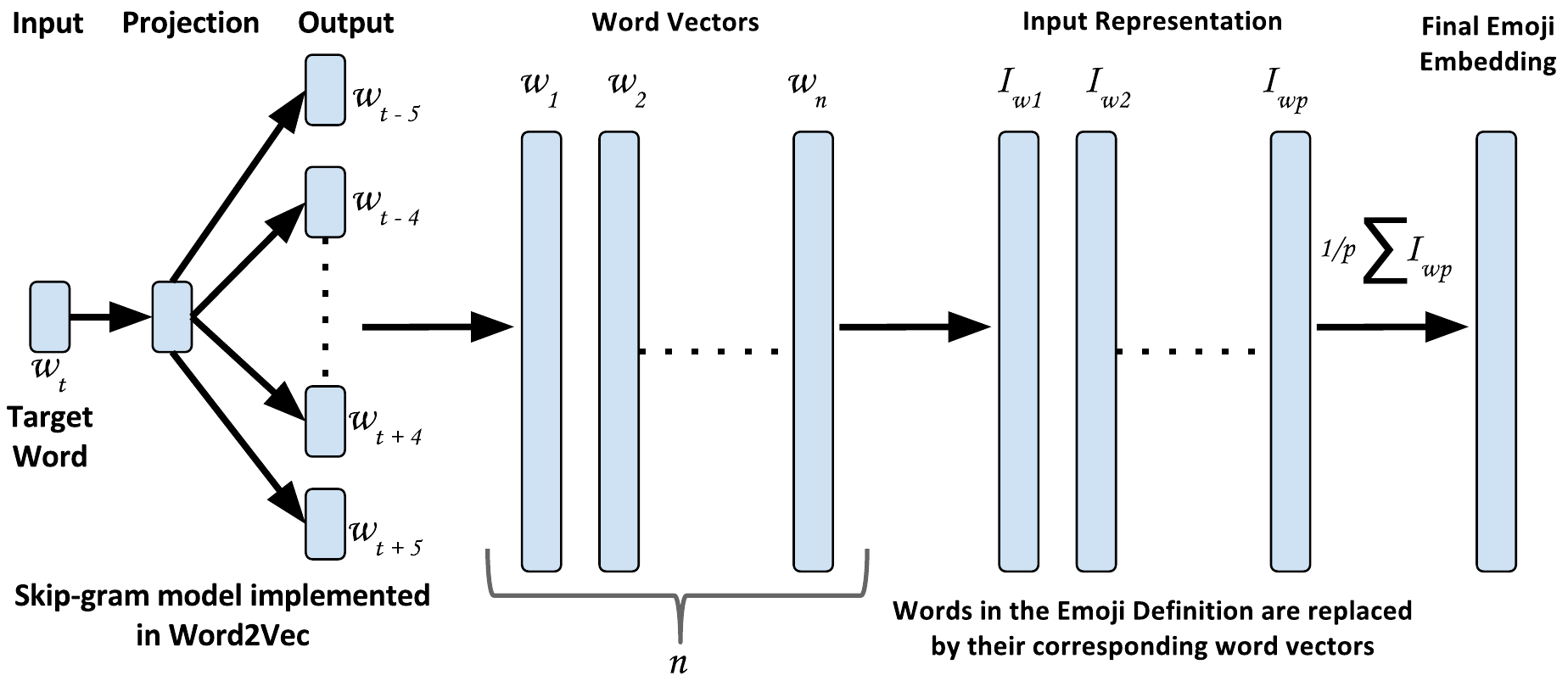} 
\caption{Learning Emoji Embedding Models using Word Vectors}
\label{fig:emoembedding}
\end{figure*}

We chose two different types of datasets, namely, a tweet corpus and a Google News corpus, to train emoji embedding models. We made this selection to make it easy to compare our emoji embedding models with other works that have used embedding models based on tweet text and Google News text. To train the Twitter word embedding model, we first collected a Twitter dataset that contained emoji using the Twitter public streaming API\footnote{\url{https://dev.twitter.com/streaming/public}}. The dataset was collected using emoji Unicodes as filtes over a four week period, from August 6$^{th}$, 2016 to September 8$^{th}$, 2016. It consists of 147 million tweets containing emoji. We first removed all retweets and then converted all emoji in the remaining 110 million unique tweets into textual features using the Emoji for Python\footnote{\url{https://pypi.python.org/pypi/emoji/}} API. The tweets were then stemmed before being processed with Word2Vec~\cite{DBLP:journals/corr/abs-1301-3781} using a Skip-gram model with negative sampling. This process is depicted in Figure~\ref{fig:emoembedding}. We choose the Skip-gram model with negative sampling to train our model as it is shown to generate robust word embedding models even when certain words are less frequent in the training corpus~\cite{NIPS2013_5021}. We set the number of dimensions of our model to 300 and the negative sampling rate to 10 sample words, which are shown to work well empirically~\cite{NIPS2013_5021}. We set the context word window to 5 (words $w_{t-5}$ to $w_{t+5}$ in Figure~\ref{fig:emoembedding}) so that it will consider 5 words to left and right of the target word (word $w_{t}$ in Figure~\ref{fig:emoembedding}) at each iteration of the training process. This setting is suitable for sentences where the average sentence length is less than 11 words, as is the case in tweets~\cite{HuTK13}. We ignore the words that occur fewer than 10 times in our Twitter dataset when training the word embedding model. We use a publicly available word embedding model that is trained over Google News corpus\footnote{\url{https://goo.gl/QaxjVC}} to obtain Google News word embeddings.

We use the learned word vectors to represent the different types of emoji definitions listed in Section~\ref{sec:emojirep}. All words in each emoji definition are replaced with their corresponding word vectors as shown in Figure~\ref{fig:emoembedding}. For example, all words in the pistol emoji's \emoji{1F52B} description, which is {\em{``A gun emoji, more precisely a pistol. A weapon that has potential to cause great harm. Displayed facing right-to-left on all platforms''}} are replaced by the word vectors learned for each word. Then, to get the emoji embedding, the word vectors of all words in the emoji definition are averaged into form a final single vector of size 300 (the dimension size). The vector mean (or average) adjusts for word embedding bias that could take place due to certain emoji definitions having considerably more words than others~\cite{kenter2016siamese}. If the total number of words in the emoji definition is $p$, the combined word 
vector $\mathbf{V_{p}}$ is calculated by: \[ \mathbf{V_{p}} = 1/p\sum_{i=0}^p \mathbf{w_{i}} \]

Using the three different emoji definitions and two types of word vectors learned over Twitter and the Google News corpora, we learn six different embeddings for each emoji. Then we integrate all words in the three types of emoji definitions into a set called {\em{(Sense\_All)}} and learn two more emoji embeddings over them by using the two types of word vectors. We take this step as prior research suggests that having more words in an emoji definition could improve the embeddings learned over them~\cite{eisner2016emoji2vec,pohl2017beyond}. We thus learn a total of 8 embeddings for emoji. The utility of each embedding as a similarity measure is discussed next.

\section{Ground Truth Data Curation} \label{sec:dc}
Once the emoji embedding vectors are learned, it is necessary to evaluate how well those represent emoji meanings. For this purpose, we create an emoji similarity dataset called `EmoSim508' that consist of 508 emoji pairs which were assigned similarity scores by ten human judges. This section discusses the development of the EmoSim508 emoji similarity dataset, which is available at \url{http://emojinet.knoesis.org/emosim508.php}.

\subsection{Emoji Pair Selection}

\begin{figure}
\centering
\includegraphics[width=1.0\linewidth]{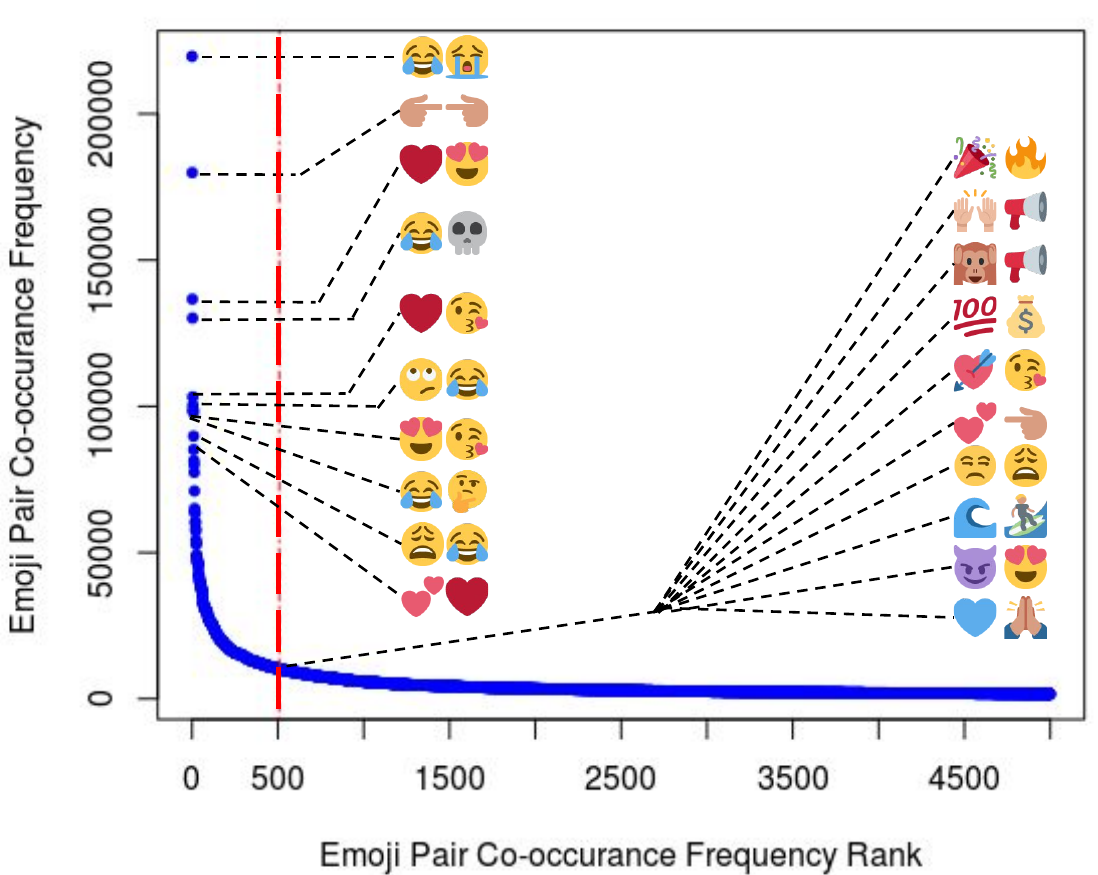} 
\caption{Emoji Co-Occurrence Frequency Graph}
\label{fig:emojigraph}
\end{figure}

Curating a reasonable sample of emoji pairs for human evaluation is a critical step: there are 2,389 emoji, leading to over 5 million emoji pairs, which would be impossible to ask a human to evaluate for their similarity. Hand-picking emoji pairs might not be a good approach as such a dataset would not cover a wide variety of similarities or could be biased towards certain relationships that commonly exist among emoji~\cite{barbieri2016does}. Furthermore, random sampling of the emoji pairs will lead to many unrelated emoji as suggested by Barbieri {\em{et al.}}~\cite{barbieri2016does}, making the dataset less useful as a gold standard dataset. We thus sought to curate the EmoSim508 dataset in such a way that the emoji pairs in it are not hand-picked but still represent a `meaningful' dataset. By meaningful, we mean that the dataset contains emoji pairs that are often seen together in practice. The dataset should also have pairs that are related, unrelated, and the shades in-between, leading to a diverse collection of examples for evaluating a similarity measure. To address this, we consider the most frequently co-occurring emoji pairs from the Twitter corpus used to learn word vectors in Section~\ref{emojiembeddinglearning} and created a plot of how often pairs of emoji co-occur with each other. From this plot, shown in Figure~\ref{fig:emojigraph}, we select the top-k emoji that cover 25\% of our Twitter dataset (shown in the dotted red line in Figure~\ref{fig:emojigraph}). This resulted in the top 508 emoji pairs. Since the co-occurence frequency plot is decidedly heavy-tailed (the blue line), we chose the 25\% threshold, giving us a dataset which is 10 times bigger than the previous dataset used by Barbieri {\em{et al.}}~\cite{barbieri2016does} to calculate emoji similarity. These 508 emoji pairs have 158 unique emoji. We have also shown the top 10 and bottom 10 emoji pairs based on their co-occurrence frequency in Figure~\ref{fig:emojigraph}. We can observe that the face emoji are dominant in the top 10 emoji pairs while bottom 10 contain few interesting emoji pairs such as \emoji{1F498} and \emoji{1F618},  \emoji{1F612} and \emoji{1F629}, and \emoji{1F30A} and \emoji{1F3C4}.

\subsection{Human Annotation Task} \label{sec:humanannotation}

We use human annotators to assign similarity scores to each emoji pair in the EmoSim508 dataset. A total of ten annotators were used, all of whom were graduate students at Wright State University, and of whom four were male and six were female. Their ages ranged from 24 years to 32 years; past studies suggest people within this age range use emoji frequently\footnote{\url{https://goo.gl/GSbCGL}}. The annotators were shown a webpage with two emoji and were prompted with two questions, one related to emoji equivalence and the other related to emoji relatedness, which they were required to answer on a five-point Lickert scale~\cite{likert1932technique} ranging from 0 to 4, where 0 means the emoji were dissimilar and 4 means the emoji were identical. We selected the five-point Lickert scale for our study for two main reasons. Firstly, past research has shown that Lickert scale is best suited for questionnaire-based user studies and five-point scale have shown to perform better than other scales (seven-points and ten-points) empirically~\cite{revilla2014choosing}. Secondly, many human annotators-involved word similarity experiments have used the same Lickert scale from 0 to 4 to calculate the similarity of words~\cite{snow2008cheap}. The two questions we asked from the annotators were:
\begin{itemize} 
\item {\bf Q1.} How equivalent are these two emoji? \\(i.e., can the use of one emoji be replaced by the other?)
\item {\bf Q2.} How related are these two emoji? \\(i.e., can one use either emoji in the same context?)
\end{itemize}
We asked Q1 to understand whether an equivalence relationship exists between an emoji pair and Q2, to understand whether a relatedness relationship exists between them. Annotators answered the same two questions for all 508 emoji pairs in the EmoSim508 dataset. We then averaged values received as answers for the ordinal selections (0 to 4) for both questions separately and assign the emoji pair an emoji equivalence score and an emoji relatedness score. Then we average the two values to calculate the final emoji similarity score for a given pair of emoji. We use the final emoji similarity score to evaluate our emoji embedding models.

\begin{table*}[]
\centering
\caption{Top-5 Emoji Pairs with Highest Inter-annotator Agreement for Each Ordinal Value from 0 to 4}
\label{annot-agreement}
\begin{adjustbox}{max width=\textwidth}
\begin{tabular}{|l|c|c|c|c|c|c|c|c|c|c|}
\hline
\multicolumn{1}{|c|}{\textbf{Ordinal Rating}}               & \multicolumn{2}{c|}{\textbf{0}} & \multicolumn{2}{c|}{\textbf{1}} & \multicolumn{2}{c|}{\textbf{2}} & \multicolumn{2}{c|}{\textbf{3}} & \multicolumn{2}{c|}{\textbf{4}} \\ \hline
\multicolumn{1}{|c|}{\textbf{Question}}             & \textbf{Q1}    & \textbf{Q2}    & \textbf{Q1}    & \textbf{Q2}    & \textbf{Q1}    & \textbf{Q2}    & \textbf{Q1}    & \textbf{Q2}    & \textbf{Q1}    & \textbf{Q2}    \\ \hline

                                                    & \emoji{1F4E2} \emoji{1F648}          & \emoji{00A9} \emoji{1F499}          & \emoji{1F64F} \emoji{1F44D}          & \emoji{1F445} \emoji{1F4A6}          & \emoji{1F60D} \emoji{1F499}          & \emoji{1F629} \emoji{1F644}          & \emoji{2764} \emoji{1F618}          & \emoji{1F48D} \emoji{1F60D}          & \emoji{1F3B5} \emoji{1F3B6}          & \emoji{1F3B5} \emoji{1F3B6}         \\ \cline{2-11}

\multicolumn{1}{|c|}{\textbf{Emoji Pairs}}                                                    & \emoji{00A9} \emoji{1F49B}             & \emoji{1F3B6} \emoji{1F525}               & \emoji{0263A} \emoji{1F618}                &  \emoji{1F631} \emoji{1F602}              & \emoji{1F602} \emoji{1F61B}               & \emoji{1F60D} \emoji{1F44C}               &  \emoji{1F449} \emoji{1F448}              & \emoji{1F649} \emoji{1F648}               & \emoji{1F495} \emoji{2764}               &  \emoji{0263A} \emoji{1F60A}              \\ \cline{2-11}

\multicolumn{1}{|c|}{\textbf{with}}    & \emoji{1F4AF} \emoji{1F649}             & \emoji{1F602} \emoji{1F525}               & \emoji{1F629} \emoji{1F644}               &  \emoji{1F644} \emoji{1F602}              & \emoji{1F3B6} \emoji{1F4FB}               & \emoji{1F61C} \emoji{1F602}                & \emoji{1F499} \emoji{1F49A}               &  \emoji{1F64A} \emoji{1F649}              & \emoji{1F38A} \emoji{1F389}               &  \emoji{1F38A} \emoji{1F389}              \\ \cline{2-11}

 \multicolumn{1}{|c|} {\textbf{Highest Agreement}}                       & \emoji{1F64T} \emoji{1F618}             &  \emoji{1F4A6} \emoji{1F449}              &  \emoji{1F60A} \emoji{1F604}              & \emoji{1F918} \emoji{1F602}               & \emoji{1F602} \emoji{1F606}               & \emoji{1F525} \emoji{1F4A5}               & \emoji{1F49C} \emoji{1F618}               & \emoji{1F498} \emoji{1F60D}               & \emoji{1F495} \emoji{1F60D}               &  \emoji{2764} \emoji{1F49E}              \\ \cline{2-11}

                                                    & \emoji{1F608} \emoji{1F459}             &  \emoji{1F334} \emoji{1F60D}              &  \emoji{1F914} \emoji{1F62D}              &   \emoji{1F643} \emoji{1F602}             & \emoji{1F499} \emoji{1F618}               &  \emoji{1F389} \emoji{1F496}               &  \emoji{1F496} \emoji{1F618}             & \emoji{1F3C6} \emoji{26BD}               & \emoji{0263A} \emoji{1F60A}               &   \emoji{1F499} \emoji{1F49A}             \\ \hline
                                                    
\end{tabular}
\end{adjustbox}                                        
\end{table*}

\subsection{Annotation evaluation}

We conducted a series of statistical tests to verify that EmoSim508 is a reliable dataset, that is, to ensure that the annotators did not randomly answer the task's questions~\cite{artstein2008inter}. To verify this, we measured the inter-annotator agreement. Since we had ten annotators who used ordinal data to evaluate the similarity of emoji, we selected Krippendorff's alpha coefficient $\alpha$ to calculate the agreement among annotators~\cite{hayes2007answering}. We calculated annotator agreement for each question separately and observed an $\alpha$ value of 0.632 for Q1 and an $\alpha$ value of 0.567 for Q2. This tells us that the emoji similarity evaluation was not an easy task for the annotators and their agreement is slightly better when deciding on two emoji pairs for equivalence than relatedness. In our dataset, a lot of annotators have agreed on the non-equivalence of emoji pairs, thus, we believe that the slightly higher $\alpha$ score for agreeing on the equivalence of an emoji pair could be a result of that. 

\begin{figure}
\centering
\includegraphics[width=0.94\linewidth]{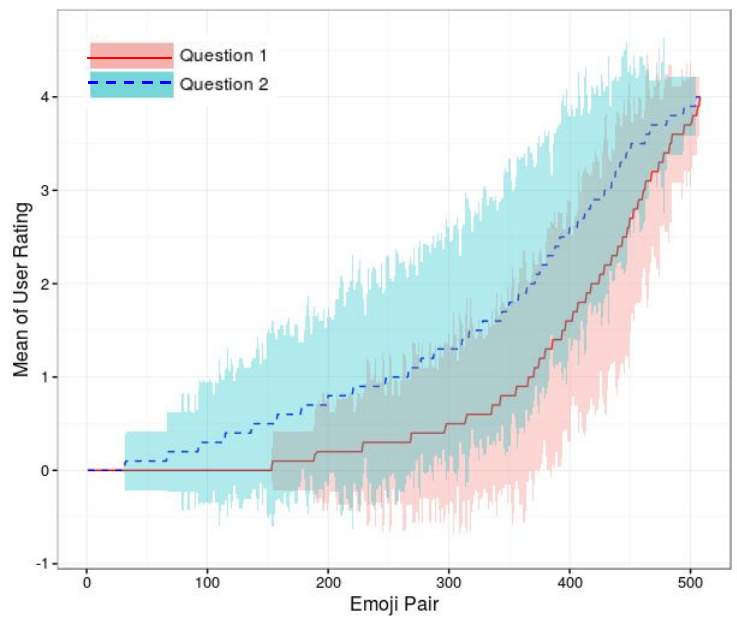} 
\caption{Distribution of the Mean of User Ratings}
\label{fig:userrating}
\end{figure}

To evaluate how reasonable are the scores provided by the human annotators, we look at the emoji pairs with highest inter-annotator agreement for each ordinal value in the Lickert scale (0 to 4) in Table~\ref{annot-agreement}. Here, we focus on annotator agreement at each level of the Lickert scale (0 to 4). We notice that all annotators have agreed that the \emoji{1F3B5} and \emoji{1F3B6} emoji show an equivalence relationship. All other emoji pairs shown for ordinal value 4, which stands for `equivalent or fully related', show high agreement (a minimum of $8/10$) among the annotators. Ordinal value 3, which stands for `highly similar or closely related', show medium agreement (a minimum of $5/10$) among annotators. Ordinal values 1 and 2, standing for `slightly similar or slightly related' and `similar or related', respectively, also show medium agreement (a minimum of $5/10$) among the top-5 emoji pairs for each ordinal value. Finally, ordinal value 0, which stands for `dissimilar or unrelated', show full agreement ($10/10$) among annotators for a total of 184 emoji pairs. The annotators have unanimously agreed that there is no relatedness and equivalence relationships exist for a group 31 and 153 emoji pairs, respectively. This further shows that it has been easier for them to agree on the dissimilarity of a pair of emoji than on its similarity or relatedness.

Figure~\ref{fig:userrating} depicts the distribution of the mean of the annotator ratings (line plot) and one standard deviation from the mean (ribbon plot) for each emoji pair for each question. For both questions, the mean of each plot shows a near-linear trend, proving that our dataset captures diverse types of relationships. Specifically, for question 1, we find a near-linear trend in the mean distribution for emoji pairs where the mean user rating is between 0.66 and 4. For question 2, we find a similar trend for emoji pairs where the mean rating is between 1 and 4. For both questions, the deviation bands are dense, especially in the range of 0.75 -- 2.5, which is to be expected. We also note that the deviation does not span beyond one rating (e.g., the deviation bands at a mean of 2 tend to span between 1 and 3). This reasonable deviation further speaks for the diversity of responses. The size of these deviation bands decrease as we approach extreme values (i.e., emoji definitely similar and definitely different). We notice an elbow (from $(0,0)$ to $(153,0)$) at the start of the mean distribution for Q1 due to the strong agreement among annotators for the unrelated emoji. This shows that even though we selected highly co-occurring emoji pairs from a Twitter corpus to be included in the EmoSim508 dataset, annotators have rated them as not related. However, we can also see that the unrelated emoji only cover 29.7\% (153/508 for Q1) of the dataset, leaving 70.3\% of the dataset with diverse relationships.

\section{Evaluating Emoji Embedding Models} \label{sec:eval} 

In this section, we discuss how we evaluated the different emoji embedding models using EmoSim508 as a gold standard dataset. We generated nine ranked lists of emoji pairs based on emoji similarity scores, one ranked list representing the EmoSim508 emoji similarity and eight ranked lists representing each emoji embedding model obtained under different corpus settings. Treating EmoSim508's emoji similarity ranks as our ground truth emoji rankings, we use Spearman's rank correlation coefficient\footnote{\url{https://goo.gl/ZA4zDP}} (Spearman's $\rho$) to evaluate how well the emoji similarity rankings generated by our emoji embedding models align with the emoji similarity rankings of the gold standard dataset. We used Spearman's $\rho$ because we noticed that our emoji annotation distribution does not follow a normal distribution. The rank correlation obtained for each setting (multiplied by 100 for display purposes) is shown in Table~\ref{spearmanranks}. Based on the rank correlation results, we notice that emoji embedding models learned over emoji descriptions moderately correlate ($40.0 < \rho < 59.0$) with the gold standard results, whereas all other models show a strong correlation ($60.0 < \rho < 79.0$). All results reported in Table~\ref{spearmanranks} are statistically significant ($p < 0.05$).

\begin{table}[]
\centering
\caption{Spearman's Rank Correlation Results}
\label{spearmanranks}
\begin{tabular}{|c|c|c|}
\hline
\textbf{Emoji Embedding Model} & \multicolumn{2}{c|}{\textbf{$\rho$ x 100 for each Corpus}}                                 \\ \cline{2-3} 
                               & \multicolumn{1}{l|}{\textbf{Google News}} & \multicolumn{1}{l|}{\textbf{Twitter}} \\ \hline
\textbf{{\em{(Sense\_Desc.)}}}                                           & 49.0                                      & 46.6                                  \\ \hline
\textbf{{\em{(Sense\_Label)}}}                                           & \textbf{76.0}                                      & \textbf{70.2}                                  \\ \hline
\textbf{{\em{(Sense\_Def.)}}}                                           & 69.5                                      & 66.9                                  \\ \hline
\textbf{{\em{(Sense\_All)}}}                                           & 71.2                                         & 67.7                                     \\ \hline
\end{tabular}
\end{table}

We observe that the emoji embeddings learned on sense labels correlate best with the emoji similarity rankings of the gold standard dataset. We further looked into what could be the reason for emoji sense labels-based embedding models {\em{(Sense\_Label)}} to outperform other models. Past work suggests that having access to lengthy emoji sense definitions could improve the performance of the emoji embedding models~\cite{eisner2016emoji2vec,pohl2017beyond}. For the 158 emoji in EmoSim508 dataset, emoji meanings were represented using 25 words on average when using the emoji descriptions; 12 words when using the emoji sense labels; 567 words when using the emoji sense definitions; and 606 words when all above definitions were combined. All our emoji embedding definitions have more words (at least twice as many) than past work~\cite{eisner2016emoji2vec}, but we notice that emoji sense labels are very specific words that only describe emoji meanings, unlike the words in emoji sense descriptions and emoji sense definitions. In contrast, emoji descriptions and emoji sense definitions provide more words describing how an emoji is shown on different platforms or how an emoji should be used in a sentence while describing the emoji's meaning. These unrelated words in emoji definitions may well be the reason for degraded performance of {\em{(Sense\_Desc.)}}, {\em{(Sense\_Def.)}} and {\em{(Sense\_All)}} embeddings. Thus, access to quality sense labels are of vital importance for learning good emoji embeddings.

\section{Emoji Embeddings at Work} \label{sec:usecase}

To show that our emoji embedding models can be used in real-world NLP tasks\footnote{Please note that our main goal is to demonstrate the usefulness of the learned embedding models and not to develop a state-of-the-art sentiment analysis algorithm.}, we set up a sentiment analysis experiment using the gold standard dataset used in~\cite{novak2015sentiment}. We selected this dataset because Barbieri {\em{et al.}}'s~\cite{barbieri2016does} and Eisner {\em{et al.}}'s~\cite{eisner2016emoji2vec} models have already been evaluated on this dataset. Thus, it allows us to compare our embedding models with theirs. The gold standard dataset consists of nearly 66,000 English tweets, labelled manually for positive, neutral or negative sentiment. The dataset is divided into a testing set that consist of 51,679 tweets, where 11,700 of them contain emoji, and a training set that consist of 12,920 tweets with 2,295 of them contain emoji. In both the training set and the test set, 46\% of tweets are labeled neutral, 29\% are labeled positive, and 25\% are labeled negative. Thus, the dataset is reasonably balanced.

\begin{table*}[]
\centering
\caption{Accuracy of the Sentiment Analysis task using Emoji Embeddings}
\label{sentimentresults}
\begin{tabular}{|l|c|c|c|c|c|c|c|c|}
\hline
\multicolumn{1}{|c|}{\textbf{}} & \multicolumn{8}{c|}{\textbf{Classification accuracy on testing dataset}}                                                                                 \\ \cline{2-9} 
\multicolumn{1}{|c|}{\textbf{Word Embedding Model}}                     & \multicolumn{2}{c|}{\textbf{N = 12,920}} & \multicolumn{2}{c|}{\textbf{N = 2,295}} & \multicolumn{2}{c|}{\textbf{N = 2,186}} & \multicolumn{2}{c|}{\textbf{N = 308}} \\ \cline{2-9} 
\multicolumn{1}{|c|}{\textbf{}}                     & \textbf{RF}         & \textbf{SVM}       & \textbf{RF}        & \textbf{SVM}       & \textbf{RF}        & \textbf{SVM}       & \textbf{RF}       & \textbf{SVM}      \\ \hline

Google News + \texttt{emoji2vec}                             & 59.5                & 60.5               & 54.4               & 59.2               & 55.0               & 59.5               & 54.5              & 55.2              \\ \hline

Google News + {\em{(Sense\_Desc.)}}                                  & 58.7                & 61.9               & 50.6                   & 55.0                  & 49.7               & 55.3               & 45.4              & 50.0              \\ \hline

Twitter + {\em{(Sense\_Desc.)}}                                      & 60.2                & 62.5               & 55.1                   & 56.7                   & 53.8               & 57.3               & 53.5              & 53.2              \\ \hline

Google News + {\em{(Sense\_Label)}}                                 & 60.3                & 63.3               & 55.0                    & \textbf{61.8}                   & 56.8               & \textbf{62.3}               & 54.2     & \textbf{59.0}     \\ \hline

Twitter + {\em{(Sense\_Label)}}                                     & \textbf{60.7}       & \textbf{63.6}      & \textbf{57.3}                    & 60.8                   & \textbf{57.5}      & 61.5      & \textbf{56.1}              & 58.4              \\ \hline

Google News + {\em{(Sense\_Def.)}}                                   & 59.0                & 62.2               & 50.3                    & 55.0                   & 51.1               & 55.2               & 48.0              & 50.6              \\ \hline

Twitter + {\em{(Sense\_Def.)}}                                       & 60.0                & 62.4               & 53.6                   & 56.2                   & 53.7               & 56.7               & 50.6              & 50.6              \\ \hline

Google News + {\em{(Sense\_All)}}                                   & 59.1                & 62.2               & 50.8                   & 55.1                   & 50.2               & 55.3               & 50.0              & 50.6              \\ \hline

Twitter + {\em{(Sense\_All)}}                                       & 60.3                & 62.4               & 53.1                   & 57.6                   & 54.1               & 56.8               & 54.5              & 50.0              \\ \hline
\end{tabular}
\end{table*}

To represent a training instance in our sentiment analysis dataset, we replaced every word in a tweet using the different embedding models learned for that word by using different text corpora. We also replaced every emoji in the tweet with its representation from a particular emoji embedding model we learned. Table~\ref{sentimentresults} shows the results we obtained for the sentiment analysis task when using different emoji embeddings. Here, Google News + {\em{(Sense\_Desc.)}} means that all words in the tweets in the gold standard dataset are replaced by their corresponding word embedding models learned by the Google News corpus and all emoji are replaced by their corresponding emoji embeddings obtained by the {\em{(Sense\_Desc.)}} model. We report classification accuracies for:~(i) the whole testing dataset (N = 12,920);~(ii) all tweets with emoji (N = 2,295);~(iii) 90\% of the most frequently used emoji in the test set (N = 2,186); and~(iv) 10\% of the least frequently used emoji in the test set (N = 308). We trained a Random Forrest (RF) classifier and a Support Vector Machine (SVM) classifier using each test data segment. We selected those two classifier models as they are commonly used for text classification problems, including the sentiment analysis experiment conducted by Eisner {\em{et al.}}~\cite{eisner2016emoji2vec} on the same gold standard dataset. Table~\ref{sentimentresults} summarizes the results obtained in the sentiment analysis task. Following Eisner {\em{et al.}}~\cite{eisner2016emoji2vec}, we also report the accuracy of the sentiment analysis task, which allows us to compare our embedding models with theirs. Accuracy is measured in settings where the testing dataset is divided into four groups based on the availability of tweets with emoji in each group. We find that the embeddings learned over emoji sense labels perform best in the sentiment analysis task, outperforming the previous best emoji embedding model~\cite{eisner2016emoji2vec} with an improvement of 7.73\%. This embedding model also yielded the best similarity ranking as per Spearman's Rank Correlation test.

As discussed in Section~\ref{sec:eval}, we believe that the inclusion of words that are highly related to emoji meanings make emoji embeddings over sense labels to learn better models to represent the meaning of an emoji, hence, outperform the other models in the sentiment analysis task. We also notice that Twitter-based emoji embedding models continue to outperform Google News-based embedding models in the majority of the test run settings. Past research on social media text processing suggests that NLP tools designed for social media text processing outperform NLP tools designed for well-formed text processing on the same task~\cite{emojineticwsm}. We believe this could be the reason why Twitter-based models continue to outperform Google News-based models. Our results, which continue to outperform Eisner {\em{et al.}}'s model~\cite{eisner2016emoji2vec}, prove that the use of emoji descriptions, sense labels, and emoji definitions to model emoji meanings has resulted in learning better emoji embedding models.

\section{Conclusion} \label{sec:con}

This paper presented how semantic similarity of emoji can be calculated by utilizing the machine-readable emoji sense definitions. Using the emoji descriptions, emoji sense labels and emoji sense definitions extracted from EmojiNet, and using two different training corpora obtained from Twitter and Google News, we explored multiple emoji embedding models to measure emoji similarity. With the help of ten human annotators who are knowledgeable about emoji, we created EmoSim508 dataset, which consist of 508 emoji pairs and used it as the gold standard to evaluate how well our emoji embedding models perform in an emoji similarity calculation task. To show a real-world use-case of the learned emoji embedding models, we used them in a sentiment analysis task and presented the results. We released the EmoSim508 dataset and our emoji embedding models with our paper. This is the first effort that explored utilizing a machine-readable emoji sense inventory and distributional semantic models to learn emoji embeddings. In the future, we would like to extend our emoji embedding models to understand the differences in emoji interpretations due to how they appear across different platforms or devices. We would also like to apply our emoji embedding models to other emoji analysis tasks such as emoji-based search. Specifically, we would like to explore whether emoji similarity results could be used to improve the recall in emoji-based search applications.

\section*{Acknowledgments}
We are thankful to the annotators who helped us in creating the EmoSim508 dataset. We acknowledge partial support from the National Institute on Drug Abuse (NIDA) Grant No. 5R01DA039454-03: ``Trending: Social Media Analysis to Monitor Cannabis and Synthetic Cannabinoid Use'', and the National Science Foundation (NSF) award: CNS-1513721: ``Context-Aware Harassment Detection on Social Media''. Points of view or opinions in this document are those of the authors and do not necessarily represent the official position or policies of the NIDA or NSF.

\bibliographystyle{splncs03}

\end{document}